# The Effects of Perfect and Sample Information on Fuzzy Utilities in Decision-Making

Maria Angeles Gil*
Computer Science Division, University of California, Berkeley

Pramod Jain
Mechanical Engineering, University of California, Berkeley

## Abstract

In this paper, we first consider a Bayesian framework and model the "utility function" in terms of fuzzy random variables. On the basis of this model, we define the "prior (fuzzy) expected utility" associated with each action, and the corresponding "posterior (fuzzy) expected utility given sample information from a random experiment". The aim of this paper is to analyze how sample information can affect the expected utility. In this way, by using some fuzzy preference relations, we conclude that sample information allows a decision maker to increase the expected utility on the average. The upper bound on the value of the expected utility is when the decision maker has perfect information. Applications of this work to the field of artificial intelligence are presented through two examples.

## 1 Introduction

In traditional decision-making problems, probabilities are numerical representations of the beliefs and the current state of information of the decision maker, whereas utilities are regarded as numerical representations of his preferences. Thus, in these problems the decision maker must be able to quantify the relative value of any situation that may arise.

In a Bayesian context, the utility function is formalized as follows (cf., [4]). Let $\Theta$ and $A$ denote the state and action spaces of the Decision problem, and let $\xi$ be the prior distribution on a measurable space defined on $\Theta$. Then, a *utility function* is a real-valued function $u$ on $\Theta \times A$ such that

- for each action $a \in A$, $u(.,a)$ is a random variable on the measurable space defined on $\Theta$, having a finite expectation with respect to $x$ denoted by $E[u(a|x)]$.

- $a$ is preferred or indifferent to $a'$ (depending on decision maker's preferences) if and only if $E[u(a|x)] \geq E[u(a'|x)]$.

The utility assessment procedures usually involve the acceptance of some conditions or axioms for the preference relations, in order to guarantee the existence of a numerical utility function (axiomatic approach to the Utility Theory). As remarked in previous papers (see, for instance, [1], [6], [8], [11], [12], [14], [17], [19], [20], and [21], the necessity for assessing utilities in terms of numerical values may be, in practice, too restrictive, whereas the use of fuzzy sets to describe utilities is often more realistic. Following the ideas of traditional decision analysis, we are going to formalize the notion of fuzzy utility function by using the concepts of fuzzy random variable and the associated expected value, as defined in [15]. On the basis of this notion, we will then establish a principle of choice among actions, in which the optimum action is that which provides the decision maker with the (prior or posterior) "highest expected utility". Since the expected utility is in this case a fuzzy number, its highest value will be determined by considering a suitable *fuzzy preference relation* based on a ranking of fuzzy numbers satisfying some desirable properties. The main contribution of this paper is the analysis of the *average worth of sample information* (from a random experiment whose distribution depends on the state in $\Theta$) and *worth of perfect information* about the state in $\Theta$ (which is seldom available in practice), for the decision maker. This enables the decision maker to conclude whether or not to perform the experiment. This comparison will also be developed through the same preference relation, and the analysis will be completed with an illustrative example, in which a particular fuzzy preference relation introduced by Kołodziejczyk [12] is considered.

The issue of worth of sample information is ubiquitous in the field of artificial intelligence. To motivate the usefulness of these concepts to any decision analysis problem we will now present two examples, one in the field of image processing and the other in medical diagnosis.

First let us consider a quality control vision system which separates acceptable components from those with defects. For the sake of simplicity let us assume that the state space consists of two states - "acceptable" and "defective" components. The prior probability distribution on the states is available and the action space consists of two states - "accept" and "reject". Suppose the manufacturing cost of the component being inspected is $x$ units and the cost associated with the time required ($t$) to inspect a component is $CT(t)$. If a good component is rejected, then the loss is $x$ units + "cost associated with late delivery" + $CT(t)$. On the other hand, if a bad component is accepted and delivered to the customer, then the cost is equal to replacement cost ($= x$ units) + "loss of reputation" + $CT(t)$. The imprecise utilities (intended as the opposite of losses) for these two cases are clearly fuzzy be-cause of the second term in the loss function.

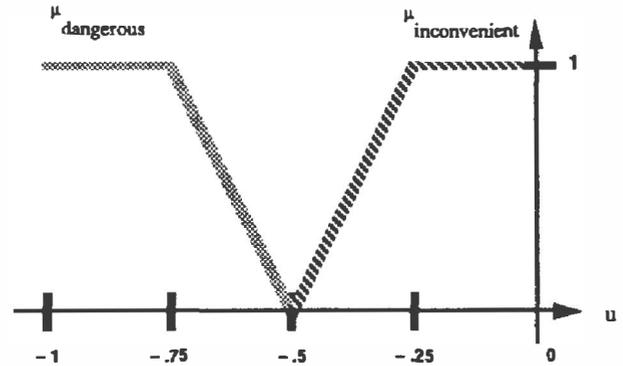

Figure 1: Membership functions of the Fuzzy Utilities "dangerous" and "inconvenient"

In a typical quality control vision system there is an abundance of information, such as the gray level of each pixel of the image. Processing of all the information is usually very expensive and time consuming, therefore one may wish to design several procedures, $P_1, \ldots, P_n$, such that procedure $P_1$ is the least accurate (w.r.t. probability of correct prediction) and procedure $P_i$ takes input from $P_{i-1}$ and predicts more accurately than $P_{i-1}$. Thus, if an accurate result is desired, then several procedures must be executed in sequence which is time consuming. The objective of this system is to maximize the utility (or minimize the loss) and since the time to inspect is also a parameter in the utility function, the worth of sample information may be used to determine when the information available is suffucient to make a "good" decision. That is, the worth of samnple information may be used as a stopping criterion for the vision algorithm.

On the other hand, in decision-making literature we can often find examples involving statistical decision problems in which utilities are quantified in terms of exact values but it would be more natural to quantify them in terms of fuzzy values. The following example have been taken from [22], an introductory text of statistics: suppose a neurologist has to classify his most serious patients as requiring exploratory brain surgery (action $a_1$) or not (action $a_2$).

From past autopsies, it has been found that 60% of the examined people needed the operation, while 40% did not. The utilities (intended as the opposite of losses) of right classifications are null. The utilities of a wrong classification are obvious: an unnecessary operation means resources are wasted and the patient may be hurt. The other utility may be worse: if a patient requiring surgery does not get it, the time lost until clear symptoms appear may be crucial. In Wonnacott and Wonnacott [22], the preceding problem is regarded as a decision problem with state space $\Theta = \{\theta_1, \theta_2\}$ ($\theta_1$ = the patient requires surgery, $\theta_2$ = the patient does not require surgery), action space $A = \{a_1, a_2\}$, and utility function $u(\theta_1, a_1) = u(\theta_2, a_2) = 0$, $u(\theta_1, a_2) = 5u(\theta_2, a_1)$, with $u(\theta_2, a_1) < 0$. However, the preceding assessment of utilities seems to be extremely precise, due to the nature of the actions and states in the problem. Thus, the following assessment could express better the decision maker (neurologist) "preferences": $\mathcal{U}(\theta_1, a_1) = \mathcal{U}(\theta_2, a_2) = 0$, $\mathcal{U}(\theta_2, a_1) =$ "inconvenient", $\mathcal{U}(\theta_1, a_2) =$ "dangerous", where $\mathcal{U}(\theta_1, a_2)$ and $\mathcal{U}(\theta_2, a_1)$ are described by means of the fuzzy sets characterized by the membership functions in Figure 1. The interest for incorporating sample information in this case is obvious: if the neurologist has to classify a serious patient as requiring or not requiring brain surgery, then he could either base his decision on the prior information or as is common in practice, try to get information regarding that patient before making a decision.

## 2 Preliminary Concepts

The following notation will be used throughout the paper. $(\Theta, \mathbf{C}, \xi)$ is a probability space associated with the state space $\Theta$, $A = \{a_1, \ldots, a_N\}$ is the set of all possible actions, and $\mathcal{F}_0(\Re)$ denote the collection of all fuzzy subsets $\tilde{V}$ of $\Re$, characterized by a membership function $\mu_{\tilde{V}} : \Re \to [0,1]$ satisfying the following properties:

(1) $supp\ \tilde{V}$ = closure of $\{w \in \Re \mid \mu_{\tilde{V}}(w) \geq 0\}$ is compact (i.e., closed and bounded).

(2) $L_\alpha(\tilde{V})$ = $\alpha$-level set of $\tilde{V}$ = $\{w \in \Re \mid \mu_{\tilde{V}}(w) \geq \alpha\}$ is closed for each $0 \leq \alpha \leq 1$.

(3) $L_1(\tilde{V})$ = modal set of $\tilde{V}$ = $\{w \in \Re \mid \mu_{\tilde{V}}(w) = 1\} \neq \emptyset$.

**Definition 2.1** *A (one-dimensional)* **fuzzy random variable** *(FRV) is a function* $\mathcal{V} : \Theta \to \mathcal{F}_0(\Re)$, *such that* $\{(\theta, w) \mid w \in L_\alpha(\mathcal{V}(\theta))\} \in \mathbf{C} \times \mathbf{B}_\Re$, *where* $\mathbf{B}_\Re$ *is the Borel $\sigma$-field on $\Re$. (In other words, the random set $L_\alpha(\mathcal{V}(.))$, defined on $\Theta$, is measurable for each $0 \leq \alpha \leq 1$).*

FRVs generalize both, random variables and random sets.

**Definition 2.2** *Let* $\mathcal{V} : \Theta \to \mathcal{F}_0(\Re)$ *be a (simple) FRV taking on the (fuzzy) values* $\tilde{V}_1, \ldots, \tilde{V}_k \in \mathcal{F}_0(\Re)$ *on* $\mathbf{C}_1, \ldots, \mathbf{C}_k \in \mathbf{C}$, *respectively (where $\bigcup_{j=1}^{k} \mathbf{C}_j = \Theta$, and $\mathbf{C}_i \bigcup \mathbf{C}_j = \emptyset$ for $i \neq j$). Then, the* **expected value** *of $\mathcal{V}$ with respect to the probability measure $\xi$ on $(\Theta, \mathbf{C})$ is the fuzzy set* $\tilde{E}(\mathcal{V}|\xi) = \int_\Theta \mathcal{V}(\theta)\ d\xi(\theta) \in \mathcal{F}_0(\Re)$ *given by*

$$\tilde{E}(\mathcal{V}|\xi) = \sum_{j=1}^{k} \tilde{V}_j\ \xi(\mathbf{C}_j) \qquad (1)$$

For a more general FRV, the definition of its expected value can be found in [15]. However, in practice simple FRV are usually sufficient to model imprecise utilities.

**Remark 2.1** In the original paper of Puri and Ralescu [15], the expected value of a FRV is introduced in a different manner - as a generalization of the Aumann integral of a random set [2]. Even though their final definition more complex it is equivalent to Definition 2.2. Thus, $\tilde{E}(\mathcal{V}|\xi)$ is defined as the unique fuzzy set with the property $L_\alpha(\tilde{E}(\mathcal{V}|\xi))$ = Aumann integral of the random set $L_\alpha(\mathcal{V}(.))$, for all $0 \leq \alpha \leq 1$.

Finally, to extend the notion of utility function to fuzzy utility function we need to consider a third element: comparison of expected values. Since the expected values are fuzzy numbers, the comparison operation reduces to ranking of fuzzy numbers. Several procedures for ranking have been proposed in the literature of fuzzy numbers. Some of them were introduced so that the calculations in the set of fuzzy numbers, with respect to the fuzzy addition and product by a positive real number, could be performed in a manner analogous to the operations on real numbers (and, consequently, the calculations through the expected value for a FRV could be performed in an analogous way as for random variables). More precisely, we can consider any suitable ranking of fuzzy numbers (generically denoted by $\succeq$) such that if $\tilde{U}, \tilde{V}, \tilde{W}$, and $\tilde{T}$ are four fuzzy numbers such that $\tilde{U} \succeq \tilde{V}$ and $\tilde{W} \succeq \tilde{T}$, then $\tilde{U} + \tilde{W} \succeq \tilde{V} + \tilde{T}$ (where $+$ = fuzzy addition), and $\lambda.\tilde{U} \succeq \lambda.\tilde{V}$ (where $\lambda.$ means the product by a positive scalar $\lambda$). On the other hand, $\succeq$ must also satisfy $\tilde{U} + (-\tilde{U}) \succeq \tilde{0}$ and $\tilde{0} \succeq \tilde{U} + (-\tilde{U})$ (where $\tilde{0}$ is the especial number assigning membership function equal to 1 to the value 0 and equal to 0 otherwise, and $(-\tilde{U})$ is the opposite to $\tilde{U}$). Since the purpose of this paper is not to discuss the best method for ranking fuzzy numbers, we will choose a ranking method from [12] which satisfies the preceding properties (Definition 2.3). The choice is only for the sake of performing computations in the illustrative example in Section 6, and is not a basic requirement for any of the subsequent analysis.

Let $\tilde{V} \in \mathcal{F}_0(\Re)$ be a fuzzy number, $\geq \tilde{V}$ and $\leq \tilde{V}$ denote the fuzzy sets of $\Re$, "more than or equal to $\tilde{V}$" and "less than or equal to $\tilde{V}$", respectively, that is

$$\mu_{\geq \tilde{V}}(w) = \begin{cases} \mu_{\tilde{V}}(w) & \text{if } w \leq z \\ 1 & \text{if } w > z \end{cases}$$

$$\mu_{\leq \tilde{V}}(w) = \begin{cases} \mu_{\tilde{V}}(w) & \text{if } w \geq z \\ 1 & \text{if } w < z \end{cases}$$

**Definition 2.3** *Let $\tilde{U}, \tilde{V} \in \mathcal{F}_0(\Re)$ be two fuzzy numbers. The coefficient*

$$R(\tilde{U}, \tilde{V}) = \frac{d_1 + d_2 + d_3}{d_4 + d_5 + 2d_3} \qquad (2)$$

*($d_1 = \mathbf{d}(\geq \tilde{U} \bigvee \geq \tilde{V}, \geq \tilde{U}), d_2 = \mathbf{d}(\leq \tilde{U} \bigvee \leq \tilde{V}, \leq \tilde{U}), d_3 = \mathbf{d}(\tilde{U} \bigcap \tilde{V}, \tilde{0}), d_4 = \mathbf{d}(\geq \tilde{U}, \geq \tilde{V}), d_5 = \mathbf{d}(\leq \tilde{U}, \leq \tilde{V})$, where $\mathbf{d}$ = Hamming distance between fuzzy sets, $\bigvee$ = extended maximum of fuzzy sets, and $\bigcap$ = intersection of fuzzy*

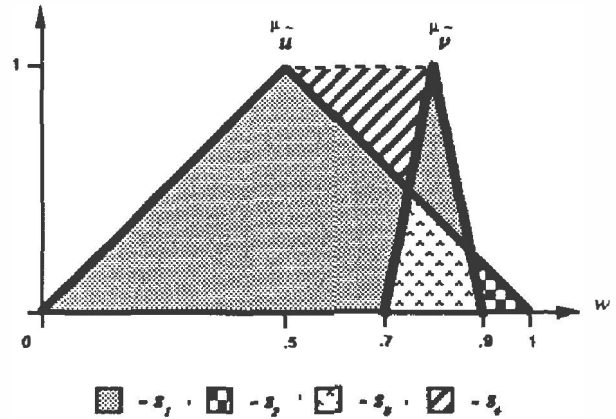

Figure 2: Areas in the second expression for Kołodiejczyk coefficient in Eq. (2)

*sets) represents the **degree of truth for the assertion "$\tilde{U}$ is not higher than $\tilde{V}$"** and*

*(1) $\tilde{U}$ is said to be **preferred or indifferent to** $\tilde{V}$, denoted $\tilde{U} \succeq \tilde{V}$, whenever $R(\tilde{U}, \tilde{V}) \leq R(\tilde{V}, \tilde{U}) = 1 - R(\tilde{U}, \tilde{V})$ (that is, whenever $R(\tilde{U}, \tilde{V}) \leq .5$).*

*(2) $\tilde{U}$ is said to be **indifferent to** $\tilde{V}$, denoted $\tilde{U} \simeq \tilde{V}$, whenever $R(\tilde{U}, \tilde{V}) = R(\tilde{V}, \tilde{U})$ (that is, whenever $R(\tilde{U}, \tilde{V}) = .5$).*

The preceding coefficient $R$ could be alternatively expressed as follows:

$$R(\tilde{U}, \tilde{V}) = \frac{S_1 + S_3 + S_4}{S_1 + S_2 + 2S_3 + 2S_4 + 2S_5}$$

where $S_1$ = areas where $\tilde{V}$ "dominates" $\tilde{U}$, $S_2$ = areas where $\tilde{U}$ "dominates" $\tilde{V}$, $S_3$ = areas where $\tilde{U}$ and $\tilde{V}$ are "indifferent", $S_4$ = areas where the greatest value of $\tilde{U}$ is lower than the smallest value of $\tilde{V}$ at the same level of membership, and $S_5$ = areas where the greatest value of $\tilde{V}$ is lower than the smallest value of $\tilde{U}$ at the same level of membership. Figure 2 explains the meaning of the areas considered in the alternative expression, and illustrates the application of coefficient $R$. Here $R(\tilde{U}, \tilde{V}) = .9$, and hence $\tilde{V}$ is preferred to $\tilde{U}$. Several examples illustrating the behavior of $R$ and an analysis of its properties may be found in [12].

# 3 Modeling the Fuzzy Utility Function

**Definition 3.1** *A fuzzy utility function is a fuzzy set-valued function $\mathcal{U}$ on $\Theta \times A$ such that*

*i) for each action $a \in A$, $\mathcal{U}(.,a)$ is a FRV on $(\Theta, \mathbf{C})$, integrable bounded, and whose expected value with respect to $\xi$ is a fuzzy number denoted by $\tilde{E}[\mathcal{U}(a|\xi)]$.*

*ii) $a$ is preferred or indifferent to $a'$ (according to the decision maker preferences) if and only if $\tilde{E}[\mathcal{U}(a|\xi)] \succeq \tilde{E}[\mathcal{U}(a'|\xi)]$.*

The assessment of fuzzy utilities has been discussed in previous studies (see, for instance, [6]).

**Remark 3.1** Conditions (1), (2) and (3) in the definition of a FRV and the assumption that the fuzzy utility function is integrable bounded (see [15]) have been imposed to guarantee that the expected value (expected utility) exists and is a fuzzy number (intended as a normalized convex fuzzy set).

**Remark 3.2** Definition 3.1 is very similar to other definitions previously considered (see, for instance [8]), but the present one is slightly more general (since it does not require the utility value to be fuzzy numbers) and is introduced by following the ideas in the traditional case.

The next result indicates that if a fuzzy utility function exists, then certain linear transformations of this function will also be utility functions. This property is analogous to a well-known result in the non-fuzzy case. Thus,

**Theorem 3.1** Let $\mathcal{U}$ be a fuzzy utility function on $\Theta \times A$. Then, the fuzzy set-valued function on $\Theta \times A$ defined by $\mathcal{V}(\theta, a) = \alpha\, \mathcal{U}(\theta, a) + \beta$ (where $\alpha > 0$ and $\beta$ are real constants) is also a fuzzy utility function.
**Proof:** Indeed, $\mathcal{V}(.,a) = \alpha\, \mathcal{U}(.,a) + \beta$ is a FRV for each action $a \in A$. Due to the properties of the rankings of fuzzy numbers we have considered, conditions i) and ii) in Definition 3.1 are both satisfied. $\triangle$

The preceding result will allow us to arbitrarily constrain $supp\ \mathcal{U}(\theta, a)$ to be contained in a particular real interval (say [0,1]), without loss of generality.

# 4 Modeling the Principle of Choice Without and With Experimentation

For any action $a \in A$, the fuzzy number $\tilde{E}[\mathcal{U}(a|\xi)]$ will be called the *prior expected utility of $a$*. In accordance with condition ii) in Definition 3.1, the existence of a fuzzy utility function entails the acceptance of the decision-making criterion based on the "maximization" of expected utility. The Bayes principle of choice may now be extended as follows:

**Definition 4.1** *An action $a^* \in A$ is called* **optimal prior action** *if $\tilde{E}[\mathcal{U}(a^*|\xi)] \succeq \tilde{E}[\mathcal{U}(a_i|\xi)]$, $i = 1, \ldots, N$.*

Generally, to increase the "highest" expected utility in a decision problem the decision maker takes advantage of the fact that additional information may reduce his uncertainty about the state in $\Theta$. In the extreme case, if he were able to get "perfect" information about this state, the problem of decision-making under uncertainty would become a problem of decision-making under certainty. Thus, if the decision maker knows for certain that the state of nature is $\theta = \theta'$, then the optimal action is the action $a(\theta') \in A$ such that $\mathcal{U}(\theta', a(\theta')) \succeq \mathcal{U}(\theta', a_i), i = 1, \ldots, N$. Nevertheless, perfect information is seldom available, and the decision maker must try to get information by performing a random experiment whose distribution depends on the state in $\Theta$.

Let $\mathbf{X}$ be a random experiment, characterized by a probability space $(X, \mathbf{B}_X, P_\theta), \theta \in \Theta$, where $X$ is a set in a Euclidean space (in most cases $\Re$), $\mathbf{B}_X$ is the smallest Borel $\sigma$-field on $X$ and $P_\theta$ is a probability measure on $(X, \mathbf{B}_X)$, so that $\theta$ is the state governing the experimental distribution. If the information obtained by performing experiment **X** is $x \in X$, then using Bayes' theorem the decision maker can use it to revise the distribution on $\Theta$ in light of the experimental information. This revision leads to the posterior distribution $\xi_x$ on $(\Theta, \mathbf{C})$, and the fuzzy number $\tilde{E}[\mathcal{U}(a|\xi_x)]$ will be called the *posterior expected utility of the action a*. The application of the decision-making criterion in Definition 4.1 allows us now to define the following:

**Definition 4.2** *An action $a_x^* \in A$, is called* **optimal posterior action given** $x$, *if* $\tilde{E}[\mathcal{U}(a_x^*|\xi_x)] \succeq \tilde{E}[\mathcal{U}(a_i|\xi_x)], i = 1, \ldots, N$.

We are now going to formalize an intuitive fact: the use of sample information entails a "gain" in expected utility on the average. Obviously, this gain will be bounded above by the "gain" in expected utility due to the use of perfect information.

## 5 Influences of Perfect and Sample Information on Expected Utility

Using the criterion in Definition 4.1, the "highest" expected utility for the decision maker, under prior information, is equal to $\tilde{E}[\mathcal{U}(a^*|\xi)]$. If perfect information is available, and $a(\theta') \in A$ is optimal under perfect information $\theta = \theta'$, then the utility is given by $\mathcal{U}(\theta', a(\theta'))$. Consequently, the value of this information when $\theta = \theta'$ could be measured by means of the fuzzy substraction $\mathcal{U}(\theta', a(\theta')) - \mathcal{U}(\theta', a^*)$, and hence the *Expected Value of Perfect Information* would be equal to the fuzzy number $EVPI = \int_\Theta \mathcal{U}(\theta', a(\theta')) \, d\xi(\theta') - \tilde{E}[\mathcal{U}(a^*|\xi)]$. If the decision maker obtains sample information $x$ by performing $\mathbf{X} = (X, \mathbf{B}_X, P_\theta), \theta \in \Theta$, the "highest" expected utility would be equal to $\tilde{E}[\mathcal{U}(a_x^*|\xi_x)]$. Thus, the *Expected Value of Sample Information* from **X**, could be measured by the fuzzy number $EVSI(\mathbf{X}) = \int_X \tilde{E}[\mathcal{U}(a_x^*|\xi_x)] \, dP(x) - \tilde{E}[\mathcal{U}(a^*|\xi)]$ (where $P(x)$ is the marginal probability measure on $(X, \mathbf{B}_X)$, obtained from $P_\theta(x)$ and $\xi(\theta)$ by applying the generalized Total Probability Rule).

We are now going to compare the three situations above in terms of the considered preference relations.

**Theorem 5.1** *Regardless of the prior distribution on* $\xi$, $EVPI \succeq EVSI(\mathbf{X}) \succeq \tilde{0}$, *whatever the random experiment* **X** *may be.*
**Proof:** Indeed, for all $x \in X$, we have

$$\int_\Theta \mathcal{U}(\theta', a(\theta')) d\xi_x(\theta') \succeq \tilde{E}[\mathcal{U}(a_x^*|\xi_x)] \succeq \tilde{E}[\mathcal{U}(a^*|\xi_x)]$$

By virtue of the properties of $\succeq$ with respect to addition of fuzzy numbers and product by a positive constant, we conclude that

$$\int_X \int_\Theta \mathcal{U}(\theta', a(\theta')) \, d\xi_x(\theta') \, dP(x) \succeq$$

$$\int_X \tilde{E}[\mathcal{U}(a_x^*|\xi_x)] \, dP(x) \succeq$$

$$\int_X \tilde{E}[\mathcal{U}(a^*|\xi_x)] \, dP(x) = \tilde{E}[\mathcal{U}(a^*|\xi)] \quad \triangle$$

**Remark 5.1** When the selection of $a_x^*$ is possible for all $x \in X$, fuzzy operations [5,7] guarantee that the $EVSI$ could be alternatively computed as follows:

$$EVSI(\mathbf{X}) = \sum_{i=1}^N \tilde{E}[\mathcal{U}(a_i|\xi_{X(a_i)})] P(X(a_i)) - \tilde{E}[\mathcal{U}(a^*|\xi)]$$

where $X(a_i) = \{x \in X | a_x^* = a_i\} \in \mathbf{B}_X$. In this alternative computation scheme, $EVSI(\mathbf{X})$ can be regarded as the expected value of a simple FRV.

## 6 Illustrative example

We will now examine the neurologist example from the introduction section to illustrate results in Theorem 5.1. If the neurologist has to classify a serious patient with no information other than the prior information, then he can obtain that $\tilde{E}[\mathcal{U}(a_1|x)] = .4\,\mathcal{U}(\theta_2, a_1), \tilde{E}[\mathcal{U}(a_2|\xi)] =$

.6 $\mathcal{U}(\theta_1, a_2)$, so that $R(\tilde{E}[\mathcal{U}(a_1|\xi)], \tilde{E}[\mathcal{U}(a_2|\xi)]) = 0$, and hence $a_1$ is preferred to $a_2$. Thus, $\tilde{E}[\mathcal{U}(a^*|\xi)] = \tilde{E}[\mathcal{U}(a_1|\xi)]$.

Suppose that the neurologist tries to make his decision on the basis of the information supplied by a combined score $\mathbf{X}$ obtained from several tests. Past experiences have shown that $\mathbf{X}$ is normally distributed with variance equal to 64 and mean equal to 120 for those who require surgery and 100 for those who do not. On the basis of the information from $\mathbf{X}$ we can revise the prior distribution on $\Theta$ to obtain the posterior ones. Then, by computing $\tilde{E}[\mathcal{U}(a|\xi_x)]$ for each $a \in A$ and $x \in X = \Re$, we conclude that it may be possible to determine $a_x^*$ in a generic way for each $x \in X$. In this example $a_x^* = a_1$ for $x \geq 110 - 3.2 \log 6.5$, $a_x^* = a_2$ otherwise. Consequently, $EVPI = -.4\,\mathcal{U}(\theta_2, a_1)$, and $EVSI(\mathbf{X}) = .1234\,\mathcal{U}(\theta_2, a_1) + .0136\,\mathcal{U}(\theta_1, a_2) - .4\,\mathcal{U}(\theta_2, a_1)$, whence $R(EVSI(\mathbf{X}), EVPI) = 0$, and $R(EVSI(\mathbf{X}), \tilde{0}) = .2973$, that is, the $EVSI(\mathbf{X})$ is not higher than the $EVPI$ and $EVSI(\mathbf{X})$ is non-negative with a high degree of truth (.7027).

## 7 Concluding Remarks

The study in this paper can be immediately extended to the case in which the prior distribution on the state space is fuzzy. In order to express the prior available information (non-sample information) in probabilistic terms, most (although not all) Bayesians follow, if necessary, the subjective interpretation of probabilities. The description of these probabilities by means of imprecise propositions (such as, "likely", "improbable", "very likely", and so on), is often more realistic than the numerical one. The decision-making problem with fuzzy probabilities and fuzzy utilities, has been examined in previous papers (see, for instance, [8], [6]). We now propose to develop a study similar to the present one by modeling fuzzy utilities through FRV, and using the arithmetic operations on fuzzy probabilities in [10]. Another immediate extension would be one in which it is assumed that fuzziness is present in sample information, along the lines of [9], [18], and [25].

Results in Section 5 suggest a more exhaustive analysis of the worth of sample information in decision-making problems with fuzzy utilities. Thus, it would be interesting to analyze the extended Expected Value of Sample Information in those problems, and to employ this fuzzy value to compare experiments and select one which provides the decision maker with the highest extended EVSI. Comparison of experiments is a well established statistical theory developed by Blackwell. He introduced a criterion, [3], based on statistical *sufficiency*, in which the purpose is to get sample information containing as much probabilistic information as possible regarding the state (without making reference to a decision-making context). It is a usual practice in this type of study to check the suitability of new criteria by examining their implications with Blackwell's. Thus, it would be also useful to analyze the connections of Blackwell's sufficiency criterion with the criterion to compare experiments based on the extended $EVSI$.

## Acknowledgements

This research is supported in part by NASA Grant NCC 2-275, AFOSR Grant 89-0084, NSF PYI Grant DMC-84511622 and a Grant from the Spanish MEC. Their financial support is gratefully acknowledged. The authors would like to thank Professor Alice M. Agogino for her invaluable discussions, suggestions and constant encouragement. We would also like to thank Professor Lotfi A. Zadeh and members of his research group for all their helpful comments.